\title{emnlp 2017 coreference camera ready}
\documentclass[11pt,letterpaper]{article}
\usepackage{emnlp2017}
\usepackage{times}
\usepackage{latexsym}
\usepackage{multirow}
\usepackage{graphicx}

\emnlpfinalcopy

\def\emnlppaperid{***}

\title{Event Coreference Resolution by Iteratively Unfolding Inter-dependencies among Events}

 \author{Prafulla Kumar Choubey \and Ruihong Huang \\
         Department of Computer Science and Engineering\\
		Texas A\&M University\\
         {\tt (prafulla.choubey, huangrh)@tamu.edu}}

\date{}

\begin{document}

\maketitle

\begin{abstract}
  We introduce a novel iterative approach for event coreference resolution that gradually builds event clusters by exploiting inter-dependencies among event mentions within the same chain as well as across event chains. Among event mentions in the same chain, we distinguish within- and cross-document event coreference links by using two distinct pairwise classifiers, trained separately to capture differences in feature distributions of within- and cross-document event clusters. Our event coreference approach alternates between WD and CD clustering and combines arguments from both event clusters after every merge, continuing till no more merge can be made. And then it performs further merging between event chains that are both closely related to a set of other chains of events. Experiments on the ECB+ corpus show that our model outperforms state-of-the-art methods in joint task of WD and CD event coreference resolution.
\end{abstract}

\section{Introduction}
Event coreference resolution is the task of identifying event mentions and clustering them such that each cluster represents a unique real world event. The capability of resolving links among co-referring event identities is vital for information aggregation and many NLP applications, including topic detection and tracking, information extraction, question answering and text summarization~\cite{humphreys1997event, allan1998topic, daniel2003sub, narayanan2004question,mayfield2009cross, zhang2015cross}. Yet, studies on event coreference are few compared to the well-studied entity coreference resolution.

Event mentions that refer to the same event can occur both within a document (WD) and across multiple documents (CD). One common practice \cite{lee2012joint} to approach CD coreference task is to resolve event coreference in a mega-document created by concatenating topic-relevant documents, which essentially does not distinguish WD and CD event links. 

However, intuitively, recognizing CD coreferent event pairs requires stricter evidence compared to WD event linking because it is riskier to link two event mentions from two distinct documents rather than the same document. In a perfect scenario where all WD event mentions are properly clustered and their participants and arguments are combined within a cluster, CD clustering can be performed with ease as sufficient evidences are collected through initial WD clustering. Therefore, another very common practice for event coreference is to first group event mentions within a document and then group WD clusters across documents \cite{yang2015hierarchical}. 

Nonetheless, WD coreference chains are equally hard to resolve. Event mentions in the same document can look very dissimilar ("killed/ VB" and "murder/ NN"), have event arguments (i.e., participants and spatio-temporal information of an event \cite{bejan2010unsupervised}) partially or entirely omitted, or appear in distinct contexts compared to their antecedent event mentions, partially to avoid repetitions. Under this irresolute state, approaching WD and CD individually is incompetent.

While CD coreference resolution is overall difficult, we observe that some CD coreferent event mentions, especially the ones that appear at the beginning of documents, share sufficient contexts and are relatively easier to resolve. At the same time, many of them bear sufficient differences that can bring in new information and further lead to more WD merges and consequently more CD merges. 

Guided by these observations, we present an event coreference approach that exploits inter-dependencies among event mentions within an event chain both within a document and across documents, by sequentially applying WD and CD merges in an alternating 
manner until no more merge can be made. We combine argument features of event mentions after each CD (WD) merge in order to resolve more difficult WD (CD) merges in the following iterations. Furthermore, our model uses two distinct pairwise classifiers that are separately trained with features intrinsic to each type. 
Specifically, the WD classifier uses features based on event mentions and their arguments while the CD classifier relies on features characterizing surrounding contexts of event mentions as well. 

We further exploit second-order inter-dependencies across event clusters in order to resolve additional WD and CD coreferent event pairs. Intuitively, if two event mentions are related to the same set of events, it is likely that the two event mentions refer to the same real world event, even when their word forms and local contexts are distinct. Specifically, we merge event clusters if their event mentions are tightly associated (i.e., having the same dependency relations) or loosely associated (i.e., co-occurring in the same sentential context) with enough (i.e., passing a threshold) other events that are known coreferent. 

Experimental results on the benchmark event coreference dataset, ECB+ \cite{cybulska2014using,cybulska2014guidelines}, show that our model extensively exploits inter-dependencies between events and outperforms the state-of-the-art methods for both WD and CD event coreference resolution.

\begin{figure*}
    \centering \includegraphics [width=6.4in]{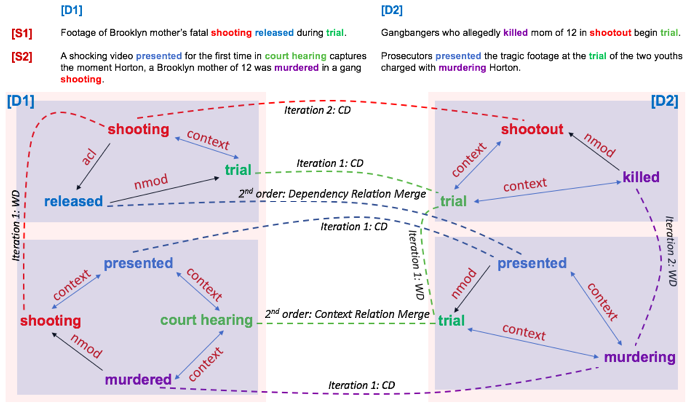}
  \caption{An example of Event Coreference using the iterative two stage model. All event mentions are boldfaced; solid arrow line between event mentions show second order relations between them; dashed lines link coreferent event mentions and are tagged with the type of merge.}
\label{examplegraph}
\end{figure*}

\section{Related Work}

Different approaches, focusing on either of WD or CD coreference chains, have been proposed for event coreference resolution.
Works specific to WD event coreference includes pairwise classifiers \cite{ahn2006stages,chen2009pairwise} graph based clustering method \cite{chen2009graph}, information propagation \cite{liu2014supervised}, and markov logic networks \citet{lujoint}.
As to only CD event coreference, \citet{cybulska2015translating} created pairwise classifiers using features indicating granularities of event slots and in another work (\citeyear{cybulska2015bag}), grouped events based on compatibilities of event contexts.

Like this work, several studies have considered both WD and CD event coreference resolution task together. However to simplify the problem, they \cite{lee2012joint, bejan2010unsupervised, bejan2014unsupervised} created a meta-document by concatenating topic-relevant documents and treated both as an identical task. Most recently, ~\citet{yang2015hierarchical} applied a two-level clustering model that first groups event mentions within a document and then groups WD clusters across documents in a joint inference process. Our approach advances these works and emphasizes on different natures of WD and CD clusters along with the benefits of distinguishing WD merges from CD merges and exploiting their mutual dependencies.

Iterative models, in general, have been applied to both entity coreference resolution \cite{singh2009bi, clarkentity, clarkimproving, wiseman2016learning} and prior event coreference resolution \cite{lee2012joint} works, which gradually build clusters and enable later merges to benefit from earlier ones. Especially, \citet{lee2012joint} used an iterative model to jointly build entity and event clusters and showed the advantages of information flow between entity and event clusters through semantic role features. Our model, by alternating between WD and CD merges, allows the multi-level flow of first order interdependencies. Moreover, additional cross cluster merges based on 2nd order interdependencies effectively exploits the semantic relations among events, in contrast to only semantic roles (between events and arguments) used in previous work. 

\section{System Overview and A Worked Example}

Inter-dependencies among event mentions can be effectively exploited by conducting sequential WD and CD merges in an iterative manner. In addition, recognizing second order relations between event chains relies on adequate number of event mentions that are already linked. Therefore, our model conducts event coreference in  \underline{two} stages. In the \underline{first} stage, it iteratively conducts WD and CD merges as suggested by pairwise WD and CD merging classifiers respectively. Argument features of individual event mentions are propagated within a cluster after each merge operation. In the \underline{second} stage, it explores second order relations across event clusters w.r.t context event mentions in order to carefully generate candidate event clusters and perform further merging.

The example in Figure \ref{examplegraph} illustrates the two stages of our proposed approach. It shows two iterations of WD and CD merges. In iteration 1, relatively easy coreferent event mentions were linked, including the two {\it shooting} and two {\it trial} event mentions in $doc_1$ and $doc_2$ as well as the event mentions {\it presented}, {\it trial} and {\it murder} across the two documents. Argument propagation was conducted after each merge and {\it murder}'s argument "mother of 12" in $doc_1$ is combined with the {\it murder} event in $doc_2$ after iteration 1. Then in iteration 2, more merges were made by recognizing additional coreferent event mentions including event mentions in one document (e.g., {\it murdering} and {\it killed} in $doc_2$) and event mentions across the two documents (e.g., {\it shooting} in $doc_1$ and {\t shootout} in $doc_2$). Next, two additional merges were made by leveraging second-order inter-dependencies. Specifically, both the event mentions {\it released} in $doc_1$ and {\it presented} in $doc_2$ are in the same dependency relation (``nmod'') with a mention of the {\it trial} event cluster, therefore, a new merge was made between clusters containing the two mentions.
Following this, the event mentions {\it court\ hearing} in $doc_1$ and {\it trial} in $doc_2$ were identified to have multiple coreferent events in their sentential contexts, therefore, the clusters containing these two event mentions were merged as well.

\section{Detailed System Description: Exploiting Interdependencies between Events}

\subsection{Document Clustering}
Our approach starts with a pre-processing step that clusters input documents ($\mathcal{D}$) into a set of document clusters ($\mathcal{C}$). This is meant to reduce search space and mitigate errors ~\cite{lee2012joint}. In our experiments, we used the Affinity Propagation algorithm~\cite{sklearn_api} on $tf-idf$ vectors, where terms are only proper nouns and verbs (excludes reporting and auxiliary verbs) in the document. While it is interesting to understand the influences of wrong document clusters to event coreference, this algorithm yielded perfect document clusters on the benchmark ECB+ dataset \cite{cybulska2014using,cybulska2014guidelines}. This is consistent with the prior study \cite{lee2012joint} on the related ECB dataset \cite{bejan2010unsupervised} \footnote{The ECB+ dataset is an extended version of the ECB dataset. Both datasets have documents for the same 43 topics.}, which shows that document clustering in the ECB dataset is trivial.

\vspace*{2ex}
\noindent\fbox{%
\parbox{0.475\textwidth}{%
\vspace*{1ex}
 \underline{{\bf Algorithm 1: }}
 \vspace*{-2ex}\\
 
\hspace*{2ex}{\bf input}: set of Documents $\mathcal{D}$ \\
\hspace*{5.8ex} Within-Document Classifier: $\Theta_{\mathcal{WD}}$ \\
\hspace*{5.8ex} Cross-Document Classifier: $\Theta_{\mathcal{CD}}$ \\
\hspace*{2ex}{\fontfamily{qcr}\selectfont \small // clusters of event mentions}\\
{\bf 1}\hspace*{2ex}$\mathcal{EM}$ = $\{\} $\\
\hspace*{2ex}{\fontfamily{qcr}\selectfont \small // clusters of Documents\\}
{\bf 2}\hspace*{2ex}$\mathcal{C}$ = ClusterDocument($\mathcal{D}$)\\
{\bf 3}\hspace*{2ex}{\bf for each} document cluster $c$ in $\mathcal{C}$ {\bf do} \\
{\bf 4}\hspace*{2ex}\hspace*{1ex}$\mathcal{EM'}$ = \{Singleton Clusters\} \\
\hspace*{3ex}{\fontfamily{qcr}\selectfont \small // Iterative WD and CD Merging \\}
{\bf 6}\hspace*{2ex}\hspace*{1ex}{\bf while}\ $iterate$ {\bf do}\\
{\bf 7}\hspace*{2ex}\hspace*{3ex}$iterate$ = ${\bf False}$ \\
{\bf 8}\hspace*{2ex}\hspace*{3ex}{\bf for each} two clusters $E_1,E_2 \in \mathcal{EM'}$ s.t. $\exists e_1  \in E_1, e_2 \in E_2$, $(e_1,e_2) \in$ a Doc, {\bf and}  score($\Theta_{\mathcal{WD}}, e_1, e_2$) \textgreater \ 0.60 {\bf do} \\
{\bf 9}\hspace*{2ex}\hspace*{9.8ex}Merge($E_1,E_2,\mathcal{EM'}$) \\
{\bf 10}\hspace*{1ex}\hspace*{9.8ex}$iterate$ = ${\bf True}$ \\
{\bf 11}\hspace*{1ex}\hspace*{3ex}{\bf if not} $iterate$ {\bf break}\\ 
{\bf 12}\hspace*{1ex}\hspace*{3ex}$iterate$ = ${\bf False}$ \\
{\bf 13}\hspace*{1ex}\hspace*{3ex}{\bf for each} two clusters $E_1,E_2 \in \mathcal{EM'}$ s.t. $\exists e_1  \in E_1, e_2 \in E_2$, $(e_1,e_2) \notin$ a Doc, {\bf and}  score($\Theta_{\mathcal{CD}}, e_1, e_2$) \textgreater \ 0.90 {\bf do} \\
{\bf 14}\hspace*{1ex}\hspace*{9.8ex}Merge($E_1,E_2,\mathcal{EM'}$) \\
{\bf 15}\hspace*{1ex}\hspace*{9.8ex}$iterate$ = ${\bf True}$ \\
\hspace*{2ex}{\fontfamily{qcr}\selectfont \small // Exploiting Second-Order Inter-dependencies Across Event Chains\\}
{\bf 16}\hspace*{1ex}\hspace*{1ex}{\bf while} $\exists$ two clusters $E_1, E_2 \in \mathcal{EM'} $ s.t. GovernorModifierRelated($E_1,E_2,\Theta_{\mathcal{CD}}$) {\bf do}\\
{\bf 17}\hspace*{1ex}\hspace*{3ex}$\mathcal{EM'}$ = Merge($E_1,E_2,\mathcal{EM'}$)\\
{\bf 18}\hspace*{1ex}\hspace*{1ex}{\bf while} $\exists$ two clusters $E_1,E_2 \in \mathcal{EM'} $ s.t. ContextSimilarity($E_1,E_2,\Theta_{\mathcal{CD}}$) {\bf do}\\
{\bf 19}\hspace*{1ex}\hspace*{3ex}$\mathcal{EM'}$ = Merge($E_1,E_2,\mathcal{EM'}$)\\
{\bf 20}\hspace*{1ex}\hspace*{1ex}$\mathcal{EM}$ = $\mathcal{EM}$ + $\mathcal{EM'}$\\
{\bf 21}\hspace*{1ex}{\bf output}: $\mathcal{EM}$
}%
}

\subsection{Iterative WD and CD Merging} 
We iteratively conduct WD merges and CD merges until no more merge can be done. We train pairwise classifiers for identifying event clusters to merge. Specifically for WD merges as indicated in lines 8-10 in Algorithm 1, we iteratively go through pairs of clusters that contain a pair of within-document event mentions, one mention from each cluster. If the similarity score between the two event mentions is above a tuned  threshold of $0.6$~\footnote{all tunings are performed on Validation dataset (topics 23-25)}, we merge the two clusters. Similarly, for CD merges described in lines 13-15 of Algorithm 1, we iteratively go through pairs of clusters that contain a pair of cross-document event mentions and merge the two clusters if the similarity score between the two event mentions is above another tuned threshold of $0.9$ \footnote{Note that these high threshold for WD- and CD- classifiers are meant to retain high precision and avoid error propagation in subsequent stages. Output from each classifier is a number bounded in [0,1]. 
}. Following each cluster pair merge, arguments are combined for the two merged clusters.

\begin{figure*}
  \centering \includegraphics [width=6in]{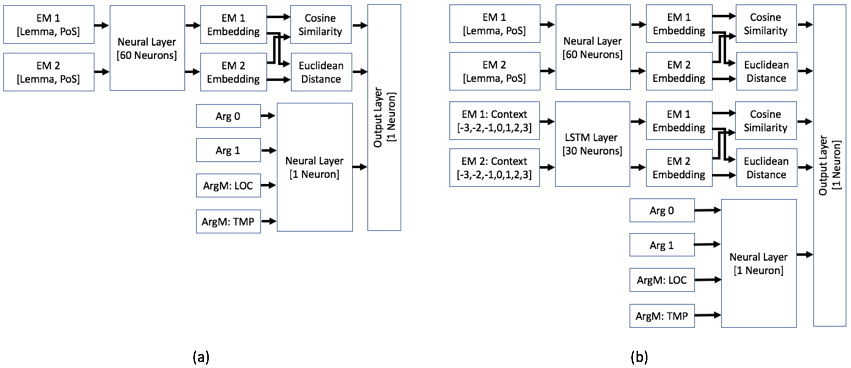}
  \caption{Pairwise	Classifiers for resolving (a) Within-Document Coreference Links (b) Cross-Document Coreference Links. {\it EM}: event mentions; {\it Arg0, Arg1, ArgM:LOC, ArgM:TMP}: semantic roles.}
  \label{classifiers-arch}
\end{figure*}

\subsection{Merging by Exploiting Second-Order Inter-dependencies Across Event Chains}
Intuitively, two event mentions that share events in their contexts are likely to be coreferent. Similarly, if their context events are coreferent, the two events are likely to be coreferent as well. 

First, if two event mentions are in the same dependency relation with two other event mentions that are known coreferent, 
then the first two event mentions are likely to describe the same real world event as well. In steps 16-17 of Algorithm 1, we perform event cluster merges by collecting evidence pertaining to dependency relations. The subroutine $GovernorModifierRelated$ ($E_1,E_2,\Theta_{\mathcal{CD}}$) checks whether two event mentions $e_1$ and $e_2$, from clusters $E_1$  and $E_2$ respectively, have a related event $e_3$ from another cluster $E_3$, such that $E_3 \notin \{E_1,E_2\}$ and pairs $(e_1, e_3), (e_2, e_3)$ are linked with the same dependency relation. 
Note that observing shared event mentions in the contexts will increase the likelihood that the two event mentions are coreferent, but we can not sufficiently infer the coreference relation yet, we still need to look at features describing the event mentions.
Therefore, if the condition was satisfied, the subroutine eventually makes merges based on the CD confidence score assigned to the event pair $(e_1, e_2)$ but using a lower threshold of $0.8$.

In addition, seeing coreferent event mentions in the sentential contexts of two events will  increase the likelihood that the two events are coreferent as well.
Then as shown in steps 18-19, we further use context events co-occurring in the same sentence as another parameter to perform additional clustering. Subroutine $ContextSimilarity$ ($E_1,E_2,\Theta_{\mathcal{CD}}$) generates a context vector ($\mathcal{CV}$) for each event cluster and check whether cosine similarity between context vectors of two clusters $E_1$ and $E_2$ ($cos(\vec{\mathcal{CV}^1},\vec{\mathcal{CV}^2}$)) is above $0.7$. Specifically, we define {\it context clusters} for an event mention as the different event clusters that have event mentions co-occurring in the same sentence. Then the context vector of an event cluster has an entry for each of its context clusters, with the value to be the number of sentences where event mentions from the two clusters co-occur. This subroutine also makes merges based on the CD confidence score using the same lower threshold of $0.8$. 

\section{Distinguishing WD and CD Merging}
We implement two distinct pairwise classifiers to effectively utilize the distributional variations in WD and CD clusters. The first classifier (WD) is used for calculating a similarity between two event mentions within a document and recognizing coreferent event mention pairs.  The second classifier (CD) is used for calculating a similarity between two event mentions across two documents and then identifying coreferent event mention pairs across documents. Both classifiers were implemented as neural nets ~\cite{chollet2015keras}.
The architectures of the two classifiers are shown in Figure \ref{classifiers-arch}. 

{\bf WD Classifier}: the neural network based WD classifier essentially inherits the features that have been shown effective in previous event coreference studies ~\cite{ahn2006stages,chen2009pairwise}, including both features for event words and features for their arguments. Specifically, the classifier includes a common neural layer shared by two event mentions to embed event lemma and parts-of-speech features. Then the classifier calculates cosine similarity and euclidean distance between two event embeddings, one per event mention. In addition, the classifier includes a neural layer component to embed event arguments that are overlapped between the two event mentions. Its output layer takes the calculated cosine similarity and euclidean distance between event mention embeddings as well as the embedding of the overlapped event arguments as input, and output a confidence score to indicate the similarity of the two event mentions.

{\bf CD Classifier}: the CD classifier mimics the WD classifier except that the CD classifier contains an additional LSTM layer ~\cite{hochreiter1997long} to embed context words. The LSTM layer is shared by both event mentions in order to calculate context word embeddings for both event mentions. Specifically, three words to each side of an event word together with the event word itself are used to calculate the context embedding for each event mention. The classifier then calculates cosine similarity and euclidean distance between two context embeddings as well. The output neural net layer will take two sets of cosine similarity and euclidean distance scores that have been calculated w.r.t. context embeddings and event word embeddings, as well as the embedding of the overlapped event arguments as input, and further calculate a confidence score indicating the similarity of two event mentions across documents.

\subsection{Characteristics of WD and CD Event Linking}
In order to further understand characteristics of within- and cross-document event linking, we trained two classifiers having the same CD classifier architecture (Figure 2(b)) but with different sets of event pairs, within-document or cross-document event pairs, then analyzed the impacts of features on each type of event linking by comparing the neural net learned weights for each feature. Table \ref{weight-comp} shows the comparisons of feature weights. 

\begin{table}[h]
\small
\begin{center}
\begin{tabular}{|l|l|l|}
\hline \bf Features & \bf WD & \bf CD   \\ \hline
Event Word Embedding: Euc & 1.017 & 0.207  \\
Event Word Embedding: Cos & 1.086 & 1.142 \\
Context Embedding: Euc & 0.038 & 0.422 \\
Context Embedding: Cos & 0.004 & 3.910 \\
Argument Embedding &  0.349 & 3.270  \\
\hline
\end{tabular}
\end{center}
\caption{\label {weight-comp} Comparisons of  Feature Weights Learned Using In-doc or Cross-doc Coreferent Event Pairs, Euc: Euclidean Distance, Cos: Cosine Similarity}
\end{table}
We can see that within-document event linking mainly relies on the euclidean distance and cosine similarity scores calculated using event word features, with a reasonable amount of weight assigned to overlapped arguments' embedding as well. However, only very small weights were assigned to the similarity and distance scores calculated using context embeddings. In contrast, in the classifier trained with cross-doc coreferent event mention pairs, the highest weight was assigned to the cosine similarity score calculated using context embeddings of two event mentions. Additionally, both the cosine similarity score calculated using event word embeddings and the overlapped argument features were assigned high weights as well. The comparisons clearly demonstrate the significantly different nature of WD and CD event coreference.

\subsection{Neural Net Classifiers and Training}
In both WD and CD classifiers, we use neural network layer with 60 neurons for embedding event word features and another layer with 1 neuron for embedding argument features. Additionally, in CD classifier, we use an LSTM layer with 30 neurons to embed context features. Dropout of 0.25 was applied to both the event word neural net layer and the context layer. We used sigmoid activation function for the dense layers and tanh activation for the LSTM layer. We used 300-dimensional word embeddings and one hot 37\footnote{Corresponding to the unique 36 POS tags based on the Stanford POS tagger \cite{toutanova03} and an additional 'padding'.} dimensional pos tag embeddings in all our experiments. Therefore, input to word embedding layer is a 337-dimensional vector and to LSTM layer is 300*7 dimensional vectors. 

We train both classifiers using the ECB+ corpus \cite{cybulska2014using, cybulska2014guidelines}. We train the WD classifier using all pairs of WD event mentions that are in an annotated event chain as positive instances and using all pairs of WD event mentions that are not in an annotated event chain as negative instances. However, there are significantly more CD coreferent event mention pairs annotated in the ECB+ corpus, therefore, we randomly sampled 70\% of all the CD coreferent event mention pairs as positive instances and randomly sampled from non-coreferent CD event mention pairs as negative instances. Specifically, number of negative instances are kept 5 times of positive instances.

Note that the pairwise classifiers will be used throughout the iterative merging stage. However, after each merge, argument propagation is conducted to enrich features for each event mention in the merged cluster and the number of arguments of an event mention will grow after several merges. In order to account for the growing number of arguments in iterative merging, we augment arguments for each event mention in training instances with arguments derived from other event mentions in the same pair. The augmenting was performed randomly for only 50\% of event mentions.

\section{Evaluation}
We perform all the experiments on the ECB+ corpus~\cite{cybulska2014using,cybulska2014guidelines}, which is an extension to the earlier  EventCorefBank (ECB)~\cite{bejan2010unsupervised} dataset. We have adopted the settings used in~\citet{yang2015hierarchical}. We divide the dataset into training set (topics 1-20), validation set (topics 21-23) and test set (topics 24-43). Table \ref{stats} shows the distribution of the corpus. 

\begin{table}[h]
\small
\begin{center}
\begin{tabular}{|l|l|l|l|l|}
\hline \bf  & \bf Train & \bf Dev & \bf Test & \bf Total \\ \hline
\#Documents & 462 & 73 & 447 & 982 \\
\#Sentences & 7,294 & 649 & 7,867 & 15,810 \\
\#Event Mentions & 3,555 & 441 & 3,290 & 7,286\\
\#CD Chains & 687 & 47 & 486 & 1,220 \\
\#WD Chains & 2,499 & 316 & 2,137 & 4,952 \\
Avg. WD chain length & 2.835 & 2.589 & 2.553 & 2.686 \\
Avg. CD chain length & 5.17 & 9.39 & 6.77 & 5.98 \\
\hline
\end{tabular}
\end{center}
\caption{\label{stats} ECB+ Corpus Statistics.}
\end{table}

\begin{table*}[h]
\small
\begin{center}
\begin{tabular}{|l|l|l|l|l|l|l|l|l|l|l|}
    \hline
    \multicolumn{11}{|c|}{\bf Cross-Document Coreference Results}\\
    \hline
\multirow{2}{*}{} &  \multicolumn{3}{l|}{$B^3$} &  \multicolumn{3}{l|}{MUC} &  \multicolumn{3}{l|}{$CEAFE_e$} &CoNLL\\ & R&P&F1&R&P&F1&R&P&F1&F1\\ \hline
    LEMMA & 39.5 & 73.9 & 51.4 & 58.1 & 78.2 & 66.7 & 58.9 & 37.5 & 46.2 & 54.8\\ \hline
    Common Classifier (WD) &  46 & 72.8 & 56.4 & 60.4 & 76.8 & 68.4 & 59.5 & 42.1 & 49.3 &58\\ 
    + 2nd Order Relations &48.8 & 72.1 & 58.2 & 61.8 & 78.9 & 69.3 & 59.3 & 44.1 & 50.6 & 59.4 \\ \hline
    Common Classifier (CD) & 44.9 & 64.7 & 53 & 66.1 & 66.4 & 66.2 & 51.9 & 46.4 & 49 & 56.1\\
    + 2nd Order Relations & 52.2 & 58.4 & 55.1 & {\bf 70.4} & 66.2 & 68.3 & 54.1 & 45.2 & 49.2 & 57.5\\\hline
    WD \& CD Classifiers & 49 &71.9&58.3&63.8&78.9&70.6&59.3&48.1&53.1&60.7\\ 
    + 2nd Order Relations {\bf (Full Model)}& {\bf 56.2} & 66.6& {\bf 61} &  67.5 & {\bf 80.4} & {\bf 73.4} & 59   & {\bf 54.2}   & {\bf 56.5} & {\bf 63.6}\\ \hline
    HDDCRP \citet{yang2015hierarchical} & 40.6 &  {\bf 78.5} & 53.5 & 67.1 & 80.3 & 73.1 & {\bf 68.9} & 38.6 & 49.5 & 58.7\\
    HDP-LEX \citet{bejan2010unsupervised} & 43.7 & 65.6 & 52.5 & 63.5 & 75.5 & 69.0 & 60.2 & 34.8 & 44.1 & 55.2\\ 
    Agglomerative \citet{chen2009pairwise} & 40.2 & 73.2 & 51.9 & 59.2 & 78.3 & 67.4 & 65.6 & 30.2 & 41.1 & 53.6\\ \hline \hline
    \multicolumn{11}{|c|}{\bf Within-Document Coreference Results}\\
    \hline
    LEMMA & 56.8 &80.9&66.7&35.9&76.2&48.8&67.4&62.9&65.1&60.2\\ \hline
    Common Classifier (WD) & 59.7 & 80.5 & 68.6 & 44.6 & 75 & 55.9 & 68.2 & 67.7 & 67.9 & 64.2\\
    + 2nd Order Relations  & 62.7 & 79.4 & 70 & 50.3 & 75.2 & 60.3 & 68.6 & 70.5 & 69.5 & 66.6\\ \hline
    Common Classifier (CD) & 65.2 & 67.1 & 66.1 & 47.6 & 53.9 & 50.5 & 69.2 & 62.1 & 65.5 & 60.7\\
    + 2nd Order Relations & 66.9 & 69.1 & 68 & 56.7 & 55.1 & 55.9 & 70.4 & 63.6 & 66.8 & 62.8\\\hline
    WD \& CD classifiers  & 63.8  & 79.9 & 70.9 & 51.6  & {\bf 75.3}& 61.2 & 68.6 & 70.5  & 69.5 & 67.2 \\ 
    + 2nd Order Relations {\bf (Full Model)}& {\bf 69.2} & 76 & 72.4 & {\bf 58.5} & 67.3  & {\bf 62.6} & 67.9& {\bf 76.1} & {\bf 71.8} & {\bf 68.9}\\ \hline
    HDDCRP \citet{yang2015hierarchical} & 67.3& {\bf 85.6} & {\bf 75.4} & 41.7   & 74.3 & 53.4 & {\bf 79.8} & 65.1 & 71.7 & 66.8\\
    HDP-LEX \citet{bejan2010unsupervised} & 67.6 & 74.7 & 71.0 & 39.1 & 50.0 & 43.9 & 71.4 & 66.2 & 68.7 & 61.2\\ 
    Agglomerative \citet{chen2009pairwise} & 67.6 & 80.7 & 73.5 & 39.2 & 61.9 & 48.0 & 76.0 & 65.6 & 70.4 & 63.9 \\ \hline
\end{tabular}
\end{center}
\caption{Within- and cross-document event coreference result on ECB+ Corpus.}
\label{results}
\end{table*}

We used event mentions identified by CRF based event extractor used in ~\citet{yang2015hierarchical} and extracted event arguments by applying state-of-the-art semantic role labeling system (SwiRL~\cite{surdeanu2007combination}). In addition, we used the Stanford parser~\cite{chen2014fast} for generating dependency relations, parts-of-speech tags and lemmas. We use pre-trained Glove vectors ~\cite{pennington2014glove}\footnote{ Trained on 840 billion tokens of Common Crawl data, http://nlp.stanford.edu/projects/glove/} for word representation and one-hot vectors for parts-of-speech tags.

We evaluate our model using four commonly adopted event coreference evaluation metrics, namely, {\bf MUC}~\cite{vilain1995model}, ${\bf B^3}$~\cite{bagga1998algorithms}, ${\bf CEAF_e}$~\cite{luo2005coreference} and {\bf CoNLL F1}~\cite{pradhan-EtAl:2014:P14-2}. We used the publicly available official implementation of revised coreference scorer ({\bf v8.01}).\footnote{ https://github.com/conll/reference-coreference-scorers}

\subsection{Baseline Systems}

We compare our iterative event coreference resolution model with five baseline systems.

LEMMA: The Lemma match baseline links event mentions within- or cross- documents which have the same lemmatized head word. It is often considered a strong baseline for this task.

HDDCRP \cite{yang2015hierarchical}: The second baseline is the supervised Hierarchical Distance Dependent Bayesian Model, the most recent event coreference system evaluated on the same ECB+ dataset. This model uses distances between event mentions, generated using a feature-rich learnable distance function, as Bayesian priors for single pass non-parametric clustering. 

HDP-LEX\footnote{\label{model:yang} The results were taken from the paper \citet{yang2015hierarchical}.}: A reimplementation of the unsupervised hierarchical bayesian model by \citet{bejan2010unsupervised, bejan2014unsupervised}.

Agglomerative\textsuperscript{~\ref{model:yang}}: A Reimplementation of two-step agglomerative clustering model, WD clustering followed by CD clustering ~\cite{chen2009pairwise}. 

We have trained our systems using the same ECB+ dataset and the same set of event mentions as these prior systems. 

\subsection{Our Systems}
We evaluate several variation systems of our proposed model.

Common Classifier (WD or CD): the system implementing only the first stage of iterative WD \& CD merging. In addition, the same neural net classifier with the architecture as shown in Figure 2(a) (the WD classifier) or in Figure 2(b) (the CD classifier) was applied for both WD and CD merging.
The neural net classifiers were trained using all coreferent event mention pairs including both within-document and cross-document ones. 

WD and CD Classifiers:  distinguishes WD from CD merges by using two distinct classifiers (Figure 2(a), 2(b)) in the first stage of the algorithm.

+ 2nd Order Relations: 
after iterative WD and CD merges within each individual chain as suggested by pairwise classifiers (the first stage), further merges (the second stage) were conducted leveraging second order event inter-dependencies across event chains.

\subsection{Results}
Table \ref{results} shows the comparison results for both within-document and cross-document event coreference resolution.
In the first stage of iterative merging, using two distinct WD and CD classifiers for corresponding WD and CD merges yields clear improvements for both WD and CD event coreference resolution tasks, compared with using one common classifier for both types of merges. In addition, the second stage of iterative merging  further improves both WD and CD event coreference resolution performance stably by leveraging second order event inter-dependencies. The improvements are consistent when measured using various coreference resolution evaluation metrics. 

Our full model achieved more than 8\% of improvements when compared with the lemma matching baseline, using the CoNLL F1-score for both WD and CD coreference resolution tasks. Furthermore, it outperforms state-of-the-art HDDCRP model for both WD and CD event coreference resolution by 2.1\% and 4.9\% respectively.

\section{Discussion and Analysis}

\begin{table}[h]
\small
\begin{center}
\begin{tabular}{|l|l|l|l|l|}
    \hline
    \multicolumn{5}{|c|}{\bf Cross-Document Coreference Results}\\
    \hline
$F_{measure}$ & {$B^3$} &  {MUC} & {$CEAFE_e$} &CoNLL\\ \hline
    1 Iteration & 56&69.3&50.3&58.5 \\ 
    2 Iterations & 57.9 & 69.9 & 52.4 & 60.1\\ 
    3 Iterations & 58.3& 70.6&53.1&60.7\\ 
   \hline \hline
    \multicolumn{5}{|c|}{\bf Within-Document Coreference Results}\\
    \hline
    1 Iteration &  69.7&55.8&68.8&64.8\\
    2 Iterations & 70.2 & 60.3 & 69.4 & 66.6 \\
    3 Iterations & 70.9 & 61.2 & 69.5 & 67.2 \\ \hline
\end{tabular}
\end{center}
\caption{Per-iteration Performance Analysis for the First Stage of Iterative WD \& CD Merging.}
\label{analysis}
\end{table}

\noindent{\bf Stage I:} The first stage of our algorithm, iterative WD and CD merging, went for three iterations (See Table \ref{analysis}). Our analysis of merges in each iteration shows that most of the merges in the initial iteration are between event mentions with the same lemma or shared arguments. In the second and third iterations, more merges were between event mentions with synonymous lemmas or shared arguments that have been accumulated in previous iterations. Example merges between synonymous event mentions include {\it (nominate, nominations)}, {\it (die, death)}, {\it (murder, killing)}, {\it (hit, strike)}, {\it (attack, bomb)} etc.

\vspace*{1ex}
\noindent{\bf Stage II:} It is even more intriguing to discuss the clusters that were merged in stage 2 of merging, that leverages second order event interdependencies across event chains. 
We found that almost all of the 81 merges happening in the second stage are between event mentions that are quite dissimilar including {\it (take over, replace)}, {\it (unveil, announce)},  {\it (win, victory, comeback)},  {\it (downtime, problem, outage)}, {\it (cut, damage)}, {\it (spark, trigger)} etc. Most interestingly, two event pairs which are antonymous to each other, {\it (win, beat)} and  {\it (defeat, victory)}, were also correctly merged. 

\vspace*{1ex}
\noindent{\bf Errors:} while our iterative algorithm has gradually resolved coreference  relations between event mentions that are synonyms or distant by surface forms, many coreference links were overlooked and many unrelated events were wrongly predicted as coreferential. We analyzed our system's final predictions 
in order to identify the most common sources of errors. 

{\bf Missed Coreference Links:}  We found that many event mentions have few or no argument in their local context, and our event coreference resolution system often failed to link these event mentions with their coreferential mentions. For instance, in the following event mention pairs that were overlooked by the system, {\it (operations, raids), (operations, sweep), (suicide, hang), (prosecution, jail), and (participating, role)}, one or both event mentions do not have an argument in their local context. This is mainly because the base WD and CD classifiers heavily rely on features extracted from the local context of two event mentions, including event words and event arguments, in resolving the coreference relation. For these event mentions having few arguments identified, the iterative algorithm may get stuck from the beginning.

While it is a grand challenge to further resolve coreferential relations between event mentions that do not have sufficient local features, these missed coreference links easily break a long and influential event chain into several sub-chains, which makes event coreference resolution results less useful for many potential applications, such as text summarization. 

{\bf Wrongly Predicted Coreference Links:} 
The majority of this type of errors are between non-coreferent event mentions that have the same lemma. This is especially common among reporting event mentions and light verb mentions. For instance, we found that 24 non-coreferent event clusters corresponding to reporting events, e.g., {\it said, told} and {\it reported}, and 13 non-coreferent clusters corresponding to light verbs, e.g., {\it take, give} and {\it get}, were incorrectly merged by the system. 

\section{Conclusions and Future Work}

We presented a novel approach for event coreference resolution that extensively exploits  event inter-dependencies between event mentions in the same chain and event mentions across chains. The approach iteratively conducts WD and CD merges followed by further merges leveraging second order event inter-dependencies across chains. 
We further distinguish WD and CD merges using two distinct classifiers that capture differences of within- and cross-document event clusters in feature distributions.
Our system was shown effective in both WD and CD event coreference and has outperformed the previous best event coreference system in both tasks.  

Note that our approach is flexible to incorporate different strategies for conducting WD and CD merges. 
In the future, we plan to continue to investigate the distinct characteristics of WD and CD coreferent event mentions in order to further improve event coreference performance. 
Especially, we are interested in including additional discourse-level features for improving WD coreference merge performance, such as, features indicating the distance between two event mentions in a document.

\section*{Acknowledgments}

We want to thank our anonymous reviewers for providing insightful review comments.

\bibliography{emnlp2017}

\begin{thebibliography}{34}
\expandafter\ifx\csname natexlab\endcsname\relax\def\natexlab#1{#1}\fi

\bibitem[{Ahn(2006)}]{ahn2006stages}
David Ahn. 2006.
\newblock The stages of event extraction.
\newblock In \emph{Proceedings of the Workshop on Annotating and Reasoning
  about Time and Events}, pages 1--8. Association for Computational
  Linguistics.

\bibitem[{Allan et~al.(1998)Allan, Carbonell, Doddington, Yamron, and
  Yang}]{allan1998topic}
James Allan, Jaime~G Carbonell, George Doddington, Jonathan Yamron, and Yiming
  Yang. 1998.
\newblock Topic detection and tracking pilot study final report.

\bibitem[{Bagga and Baldwin(1998)}]{bagga1998algorithms}
Amit Bagga and Breck Baldwin. 1998.
\newblock Algorithms for scoring coreference chains.
\newblock In \emph{The first international conference on language resources and
  evaluation workshop on linguistics coreference}, volume~1, pages 563--566.

\bibitem[{Bejan and Harabagiu(2010)}]{bejan2010unsupervised}
Cosmin~Adrian Bejan and Sanda Harabagiu. 2010.
\newblock Unsupervised event coreference resolution with rich linguistic
  features.
\newblock In \emph{Proceedings of the 48th Annual Meeting of the Association
  for Computational Linguistics}, pages 1412--1422. Association for
  Computational Linguistics.

\bibitem[{Bejan and Harabagiu(2014)}]{bejan2014unsupervised}
Cosmin~Adrian Bejan and Sanda Harabagiu. 2014.
\newblock Unsupervised event coreference resolution.
\newblock \emph{Computational Linguistics}, 40(2):311--347.

\bibitem[{Buitinck et~al.(2013)Buitinck, Louppe, Blondel, Pedregosa, Mueller,
  Grisel, Niculae, Prettenhofer, Gramfort, Grobler, Layton, VanderPlas, Joly,
  Holt, and Varoquaux}]{sklearn_api}
Lars Buitinck, Gilles Louppe, Mathieu Blondel, Fabian Pedregosa, Andreas
  Mueller, Olivier Grisel, Vlad Niculae, Peter Prettenhofer, Alexandre
  Gramfort, Jaques Grobler, Robert Layton, Jake VanderPlas, Arnaud Joly, Brian
  Holt, and Ga{\"{e}}l Varoquaux. 2013.
\newblock {API} design for machine learning software: experiences from the
  scikit-learn project.
\newblock In \emph{ECML PKDD Workshop: Languages for Data Mining and Machine
  Learning}, pages 108--122.

\bibitem[{Chen and Manning(2014)}]{chen2014fast}
Danqi Chen and Christopher~D Manning. 2014.
\newblock A fast and accurate dependency parser using neural networks.
\newblock In \emph{EMNLP}, pages 740--750.

\bibitem[{Chen and Ji(2009)}]{chen2009graph}
Zheng Chen and Heng Ji. 2009.
\newblock Graph-based event coreference resolution.
\newblock In \emph{Proceedings of the 2009 Workshop on Graph-based Methods for
  Natural Language Processing}, pages 54--57. Association for Computational
  Linguistics.

\bibitem[{Chen et~al.(2009)Chen, Ji, and Haralick}]{chen2009pairwise}
Zheng Chen, Heng Ji, and Robert Haralick. 2009.
\newblock A pairwise event coreference model, feature impact and evaluation for
  event coreference resolution.
\newblock In \emph{Proceedings of the workshop on events in emerging text
  types}, pages 17--22. Association for Computational Linguistics.

\bibitem[{Chollet(2015)}]{chollet2015keras}
Fran\c{c}ois Chollet. 2015.
\newblock Keras.
\newblock \url{https://github.com/fchollet/keras}.

\bibitem[{Clark and Manning(2015)}]{clarkentity}
Kevin Clark and Christopher~D Manning. 2015.
\newblock Entity-centric coreference resolution with model stacking.

\bibitem[{Clark and Manning(2016)}]{clarkimproving}
Kevin Clark and Christopher~D Manning. 2016.
\newblock Improving coreference resolution by learning entity-level distributed
  representations.

\bibitem[{Cybulska and Vossen(2014{\natexlab{a}})}]{cybulska2014guidelines}
Agata Cybulska and Piek Vossen. 2014{\natexlab{a}}.
\newblock Guidelines for ecb+ annotation of events and their coreference.
\newblock Technical report, Technical report, Technical Report NWR-2014-1, VU
  University Amsterdam.

\bibitem[{Cybulska and Vossen(2014{\natexlab{b}})}]{cybulska2014using}
Agata Cybulska and Piek Vossen. 2014{\natexlab{b}}.
\newblock Using a sledgehammer to crack a nut? lexical diversity and event
  coreference resolution.
\newblock In \emph{LREC}, pages 4545--4552.

\bibitem[{Cybulska and Vossen(2015{\natexlab{a}})}]{cybulska2015translating}
Agata Cybulska and Piek Vossen. 2015{\natexlab{a}}.
\newblock Translating granularity of event slots into features for event
  coreference resolution.
\newblock In \emph{Proceedings of the 3rd Workshop on EVENTS at the NAACL-HLT},
  pages 1--10.

\bibitem[{Cybulska and Vossen(2015{\natexlab{b}})}]{cybulska2015bag}
Agata Cybulska and Piek Vossen. 2015{\natexlab{b}}.
\newblock “bag of events” approach to event coreference resolution.
  supervised classification of event templates.
\newblock \emph{IJCLA}, page~11.

\bibitem[{Daniel et~al.(2003)Daniel, Radev, and Allison}]{daniel2003sub}
Naomi Daniel, Dragomir Radev, and Timothy Allison. 2003.
\newblock Sub-event based multi-document summarization.
\newblock In \emph{Proceedings of the HLT-NAACL 03 on Text summarization
  workshop-Volume 5}, pages 9--16. Association for Computational Linguistics.

\bibitem[{Hochreiter and Schmidhuber(1997)}]{hochreiter1997long}
Sepp Hochreiter and J{\"u}rgen Schmidhuber. 1997.
\newblock Long short-term memory.
\newblock \emph{Neural computation}, 9(8):1735--1780.

\bibitem[{Humphreys et~al.(1997)Humphreys, Gaizauskas, and
  Azzam}]{humphreys1997event}
Kevin Humphreys, Robert Gaizauskas, and Saliha Azzam. 1997.
\newblock Event coreference for information extraction.
\newblock In \emph{Proceedings of a Workshop on Operational Factors in
  Practical, Robust Anaphora Resolution for Unrestricted Texts}, pages 75--81.
  Association for Computational Linguistics.

\bibitem[{Lee et~al.(2012)Lee, Recasens, Chang, Surdeanu, and
  Jurafsky}]{lee2012joint}
Heeyoung Lee, Marta Recasens, Angel Chang, Mihai Surdeanu, and Dan Jurafsky.
  2012.
\newblock Joint entity and event coreference resolution across documents.
\newblock In \emph{Proceedings of the 2012 Joint Conference on Empirical
  Methods in Natural Language Processing and Computational Natural Language
  Learning}, pages 489--500. Association for Computational Linguistics.

\bibitem[{Liu et~al.(2014)Liu, Araki, Hovy, and Mitamura}]{liu2014supervised}
Zhengzhong Liu, Jun Araki, Eduard~H Hovy, and Teruko Mitamura. 2014.
\newblock Supervised within-document event coreference using information
  propagation.
\newblock In \emph{LREC}, pages 4539--4544.

\bibitem[{Lu et~al.(2016)Lu, Venugopal, Gogate, and Ng}]{lujoint}
Jing Lu, Deepak Venugopal, Vibhav Gogate, and Vincent Ng. 2016.
\newblock Joint inference for event coreference resolution.

\bibitem[{Luo(2005)}]{luo2005coreference}
Xiaoqiang Luo. 2005.
\newblock On coreference resolution performance metrics.
\newblock In \emph{Proceedings of the conference on Human Language Technology
  and Empirical Methods in Natural Language Processing}, pages 25--32.
  Association for Computational Linguistics.

\bibitem[{Mayfield et~al.(2009)Mayfield, Alexander, Dorr, Eisner, Elsayed,
  Finin, Fink, Freedman, Garera, McNamee et~al.}]{mayfield2009cross}
James Mayfield, David Alexander, Bonnie~J Dorr, Jason Eisner, Tamer Elsayed,
  Tim Finin, Clayton Fink, Marjorie Freedman, Nikesh Garera, Paul McNamee,
  et~al. 2009.
\newblock Cross-document coreference resolution: A key technology for learning
  by reading.
\newblock In \emph{AAAI Spring Symposium: Learning by Reading and Learning to
  Read}, volume~9, pages 65--70.

\bibitem[{Narayanan and Harabagiu(2004)}]{narayanan2004question}
Srini Narayanan and Sanda Harabagiu. 2004.
\newblock Question answering based on semantic structures.
\newblock In \emph{Proceedings of the 20th international conference on
  Computational Linguistics}, page 693. Association for Computational
  Linguistics.

\bibitem[{Pennington et~al.(2014)Pennington, Socher, and
  Manning}]{pennington2014glove}
Jeffrey Pennington, Richard Socher, and Christopher~D Manning. 2014.
\newblock Glove: Global vectors for word representation.
\newblock In \emph{EMNLP}, volume~14, pages 1532--1543.

\bibitem[{Pradhan et~al.(2014)Pradhan, Luo, Recasens, Hovy, Ng, and
  Strube}]{pradhan-EtAl:2014:P14-2}
Sameer Pradhan, Xiaoqiang Luo, Marta Recasens, Eduard Hovy, Vincent Ng, and
  Michael Strube. 2014.
\newblock \href {http://www.aclweb.org/anthology/P14-2006} {Scoring coreference
  partitions of predicted mentions: A reference implementation}.
\newblock In \emph{Proceedings of the 52nd Annual Meeting of the Association
  for Computational Linguistics (Volume 2: Short Papers)}, pages 30--35,
  Baltimore, Maryland. Association for Computational Linguistics.

\bibitem[{Singh et~al.(2009)Singh, Schultz, and McCallum}]{singh2009bi}
Sameer Singh, Karl Schultz, and Andrew McCallum. 2009.
\newblock Bi-directional joint inference for entity resolution and segmentation
  using imperatively-defined factor graphs.
\newblock \emph{Machine Learning and Knowledge Discovery in Databases}, pages
  414--429.

\bibitem[{Surdeanu et~al.(2007)Surdeanu, M{\`a}rquez, Carreras, and
  Comas}]{surdeanu2007combination}
Mihai Surdeanu, Llu{\'\i}s M{\`a}rquez, Xavier Carreras, and Pere~R Comas.
  2007.
\newblock Combination strategies for semantic role labeling.
\newblock \emph{Journal of Artificial Intelligence Research}, 29:105--151.

\bibitem[{Toutanova et~al.(2003)Toutanova, Klein, Manning, and
  Singer}]{toutanova03}
K.~Toutanova, D.~Klein, C.~Manning, and Y.~Singer. 2003.
\newblock {Feature-Rich Part-of-Speech Tagging with a Cyclic Dependency
  Network}.
\newblock In \emph{{Proceedings of HLT-NAACL 2003}}.

\bibitem[{Vilain et~al.(1995)Vilain, Burger, Aberdeen, Connolly, and
  Hirschman}]{vilain1995model}
Marc Vilain, John Burger, John Aberdeen, Dennis Connolly, and Lynette
  Hirschman. 1995.
\newblock A model-theoretic coreference scoring scheme.
\newblock In \emph{Proceedings of the 6th conference on Message understanding},
  pages 45--52. Association for Computational Linguistics.

\bibitem[{Wiseman et~al.(2016)Wiseman, Rush, and Shieber}]{wiseman2016learning}
Sam Wiseman, Alexander~M Rush, and Stuart~M Shieber. 2016.
\newblock Learning global features for coreference resolution.
\newblock In \emph{Proceedings of NAACL-HLT}, pages 994--1004.

\bibitem[{Yang et~al.(2015)Yang, Cardie, and Frazier}]{yang2015hierarchical}
Bishan Yang, Claire Cardie, and Peter Frazier. 2015.
\newblock A hierarchical distance-dependent bayesian model for event
  coreference resolution.
\newblock \emph{Transactions of the Association for Computational Linguistics},
  3:517--528.

\bibitem[{Zhang et~al.(2015)Zhang, Li, Ji, and Chang}]{zhang2015cross}
Tongtao Zhang, Hongzhi Li, Heng Ji, and Shih-Fu Chang. 2015.
\newblock Cross-document event coreference resolution based on cross-media
  features.
\newblock In \emph{EMNLP}, pages 201--206.

\end{thebibliography}
\bibliographystyle{emnlp_natbib}

\end{document}